\documentclass[conference]{IEEEtran}
\IEEEoverridecommandlockouts
\usepackage{cite}
\usepackage{caption}
\usepackage{subcaption}
\usepackage{url}
\usepackage{amsmath,amssymb,amsfonts}
\usepackage{algorithmic}
\usepackage{graphicx}
\usepackage{textcomp}
\usepackage{xcolor}
\usepackage{array}
\usepackage{makecell}
\usepackage{csquotes}
\usepackage[pscoord]{eso-pic}
\def\BibTeX{{\rm B\kern-.05em{\sc i\kern-.025em b}\kern-.08em
    T\kern-.1667em\lower.7ex\hbox{E}\kern-.125emX}}

\begin{document}



\newcommand{\placetextbox}[3]{
	\setbox0=\hbox{#3}
	\AddToShipoutPictureFG*{ \put(\LenToUnit{#1\paperwidth},\LenToUnit{#2\paperheight}){\vtop{{\null}\makebox[0pt][c]{#3}}}
	}
}
\placetextbox{.50}{0.055}{\small{978-1-7281-2435-3/19/\$31.00 \copyright 2019 IEEE}}

\title{Low Cost 3D Printing for Rapid Prototyping and its Application}
\author{\IEEEauthorblockN{ Taha Hasan Masood Siddique\IEEEauthorrefmark{1},  Iqra Sami\IEEEauthorrefmark{2}, Malik Zohaib Nisar\IEEEauthorrefmark{3},
Mashal Naeem\IEEEauthorrefmark{4},\\ Abid Karim\IEEEauthorrefmark{5} and Muhammad Usman\IEEEauthorrefmark{1}}
\\
\IEEEauthorblockA{\textit{
\IEEEauthorrefmark{1}Faculty of Engineering Science and Technology (FEST), Iqra University, Defence View, Karachi-75500, Pakistan.} \\
Email: $$\{tahahasan,musman\}$$@iqra.edu.pk}
\textit{\IEEEauthorrefmark{2}Pakistan Navy Engineering College, Habib Rehmatullah Road, Karachi-75350, Pakistan.}\\
Email: iqrasami26@gmail.com\\

\textit{\IEEEauthorrefmark{3}NED University of Engineering and Technology, Karachi-75270, Pakistan.}\\
Email: zohaibmalik90@gmail.com\\

\textit{\IEEEauthorrefmark{4}China Machinery Engineering Corporation, Defence Housing Authority, Karachi-75500, Pakistan.}\\
Email: mashailali@yahoo.com\\

\textit{\IEEEauthorrefmark{5}Usman Institute of Technology, Karachi-75300, Pakistan.}\\
Email: akarim@uit.edu

}

\maketitle

\begin{abstract}
In the recent years of industrial revolution, 3D printing has shown to grow as an expanding field of new applications. The low cost solutions and short time to market makes it a favorable candidate to be utilized in the dynamic fields of engineering. Additive printing has the vast range of applications in many fields. This study presents the wide range of applications of the 3D printers along with the comparison of the additive printing with the traditional manufacturing methods have been shown. A tutorial is presented explaining the steps involved in the prototype printing using Rhinoceros 3D and Simplify 3D software including the detailed specifications of the end products that were printed using the Delta 3D printer.
\\
\end{abstract}


\section{Introduction}
Three-dimensional (3D) printing also formally known as Additive Manufacturing (AM) or rapid prototyping is the process of making three dimensional objects from 3D modeled data in progressive layers \cite{Paper1} \cite{Paper3} and it is currently being promoted as the spark of new industrial revolution \cite{weller2015economic}. This technique was described by Charles Hull in 1986 \cite{Paper1}. These days there is an increased demand of customization in the global markets as the industries are expected to deliver innovative design that are also low on volume production \cite{sisca2016additive}\cite{sisca2015novel} \cite{thymianidis2012modern}\cite{mellor2014additive} and 3D printing has emerged as a powerful tool in engineering for prototyping.  From fundamental aspects the elements like fabrication speed, type and quality of material and resolution of 3D printing methods must be examined for each specific application. The technology of 3D printing is being effectively implemented in various fields including robotics, medical, aerospace, automotive, jewelry, art design, architecture, fashion, food and entertainment. 3D printing received worldwide attention as an industrial revolution. It has proved to produce low budget products using advanced software and utilizing automated manufacturing process. According to the report published by Allied marketing research  3D printing is fastest growing market these days and will generate \$8.6 billion by 2020 \cite{websitecite2}. According to the another report published by Gartner \cite{Paper4}, 75\% growth in global shipment of 3D printers were highlighted for the year 2014 and it is set to double every year. 

3D printing technology is getting more popular day by day for producing objects which were difficult to produce by using traditional methods \cite{Paper9}. 3D printing has obtained extensive success over traditional methods in prototyping as well as in manufacturing of working products because of several advantages including fabrication of complex geometry with high precision, flexibility in design, material saving, short time to market, waste reducing and personal customization \cite{Paper7}\cite{Paper11} \cite{ngo2018additive}. 3D printing adoption has rapidly increased in the recent years. It has become more useful as the technology continues to advance in versatility and capabilities \cite{schniederjans2017adoption}. Commercial 3D printers have been utilized in many fields due to the cheap manufacturing, quick production  and user-friendly interface. These machines allow automatic production of objects from computer-aided design (CAD) to the final product with minimum human efforts \cite{ding2016advanced}. Keeping in view the latest advancements and adaptation of additive manufacturing, in this study we tend to highlight the applications of such printers and their utilization in industry. A brief overview of the steps involved in designing a model using simulation tool and printing a product has also been discussed in later sections. 

This paper is organized as follows: Workflow of 3D printing process is described in Section \ref{Sec2}, simulation process is discussed in Section \ref{Sec3}, discussion of methods and material is presented in Section \ref{Sec4}, results are given in Section \ref{Sec5} and comparison with traditional manufacturing is discussed in Section \ref{Sec6}. Finally, Section \ref{Sec7} discusses the entire work in general, and gives some conclusions.

\section{Workflow of 3D printing process}
\label{Sec2}
\begin{figure}[h!]
\begin{center}
\includegraphics[scale=0.25]{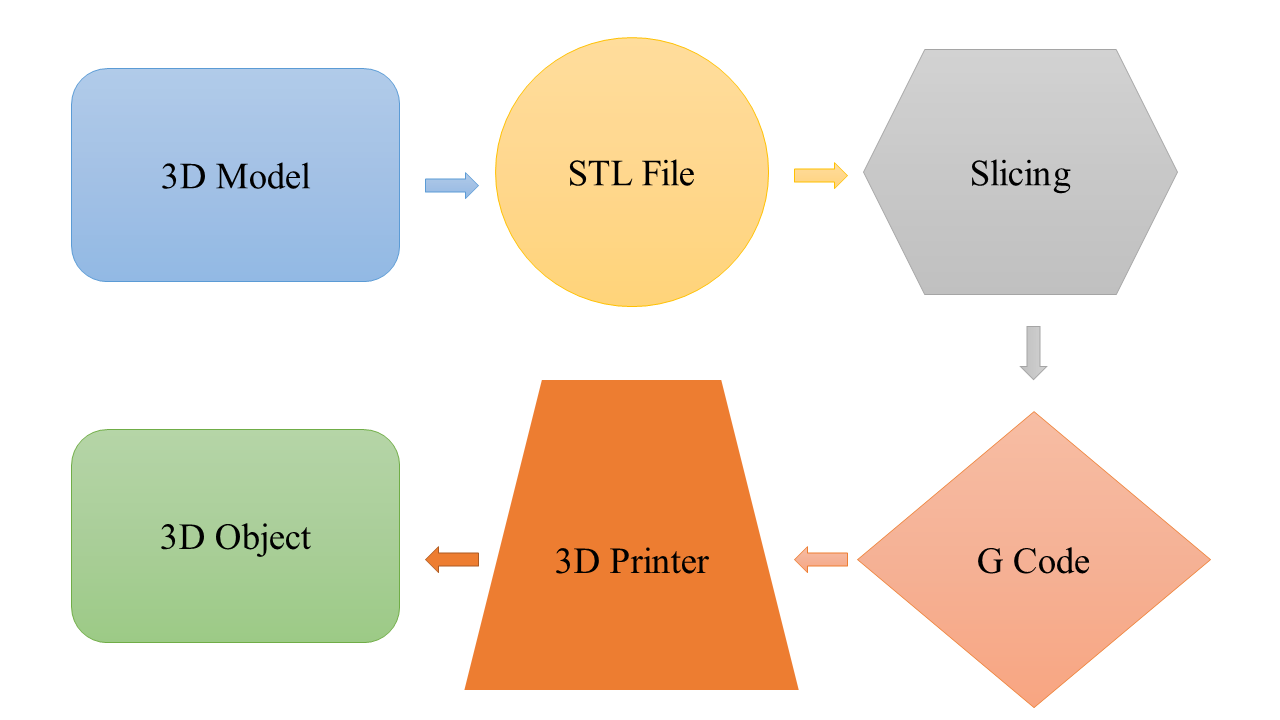}
\end{center}

\caption{Workflow of 3D Printing Process}
\label{fig1}
\end{figure}

3D printing requires an input CAD of the object which may be designed in any 3D modeling software such as Rhinoceros 3D or obtained from reverse engineering such as 3D scanners. The 3D printing process starts with a 3D model and then it is converted into the stereolithography (STL) file format which is commonly used for 3D printing \cite{onyeako2016resolution}. The process of calculating the printing path based on three dimensional model file is known as slicing. This is done by a slicing tool such as Simplify 3D. Simplify 3D creates virtual environment for the verification of 3D printing processes. This tool is helpful for previewing the printing results and preventing errors. To calculate the printing path, slicing tool tries to find out the solid shapes of the model. Therefore, the solid shapes are cut into layers and the STL file is sliced into successive layers which is favorable for printing instruction and to create extrusion paths which are filled with the plastic material and then it is converted into G-Codes \cite{baumann2016influence}. The G-Codes are easily readable for 3D printers to initiate the printing points. Printing parameters like infill percentage, nozzle diameter, extrusion multiplier, top and bottom layers and outline direction that affects the printing quality are also set by G-codes. Incorrect configurations such as inappropriate extruder settings, retraction distance, retraction speed, layer settings, infill settings and temperature settings may produce disastrous results \cite{Paper1} \cite{Paper9}.  Figure \ref{fig1} depicts a typical workflow of 3D printing process.

\section{simulation process}
\label{Sec3}
To create a 3D model, two software are generally required; a 3D modeling software such as Rhinoceros 3D that transforms the desired model into 3D model CAD file, and 3D slicing software such as Simplify 3D that divides the 3D model CAD file into thin data slices which is suitable for 3D printing. 

\subsection{3D Modeling Software}
In this study \enquote{Rhinoceros 3D} is used for 3D modeling. Rhinoceros 3D is commonly used because of its user friendly interface and large number of available tools \cite{mcneel2009rhinoceros}. The distinguishing feature of Rhinoceros 3D is that it can highlight the area of interest, remove undesired areas and export the 3D model as a CAD file in a universally acceptable stereolithography (STL) file format.

\subsection{3D Slicing Software}
\begin{figure}[ht] 
\begin{subfigure}{.20\textwidth}
\centering
\includegraphics[width=1.1\linewidth]{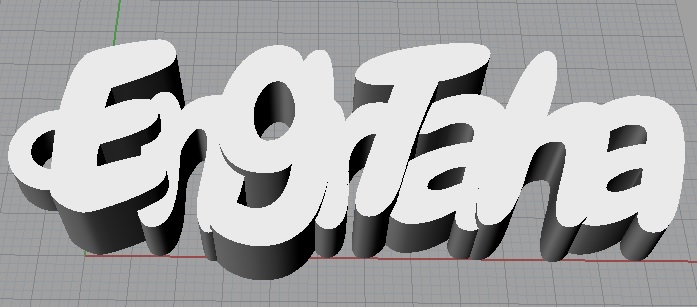}
\caption{3D Model}
\label{fig: sfig01}
\end{subfigure} 
\qquad 
\begin{subfigure}{.20\textwidth}
\centering
\includegraphics[width=1.1\linewidth]{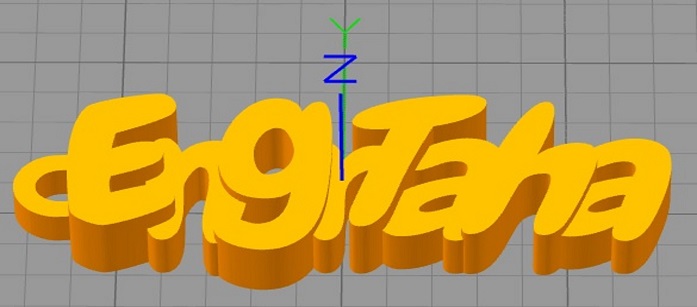}
\caption{Exported STL File}
\label{fig: sfig02}
\end{subfigure}

\begin{subfigure}{.20\textwidth}
\centering
\includegraphics[width=1.1\linewidth]{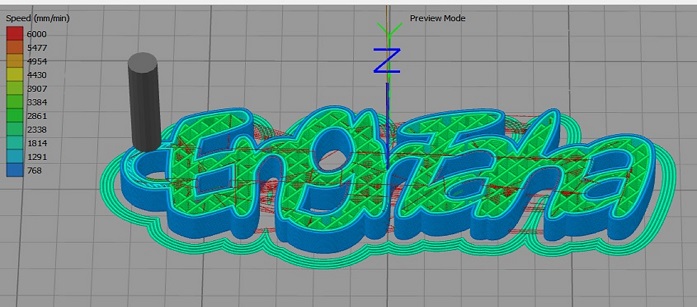}
\caption{Internal Structure}
\label{fig: sfig03}
\end{subfigure}
\qquad
\begin{subfigure}{.20\textwidth}
\centering
\includegraphics[width=1.1\linewidth]{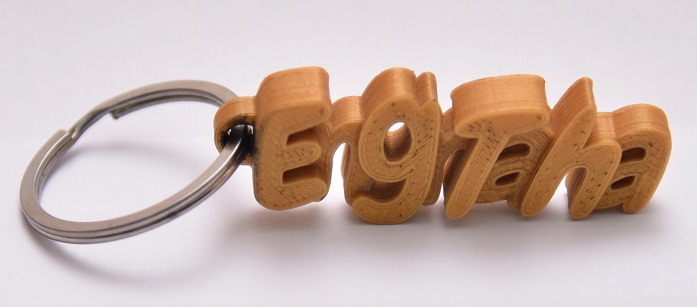}
\caption{3D Printed Product}
\label{fig: sfig04}
\end{subfigure}

\caption{3D Model to 3D Printed Product}
\label{fig2}

\end{figure}

Simplify 3D digitally slices a 3D CAD file into layers which are suitable for 3D printing. This is helpful for the orientation of the 3D CAD file related to the 3D printer to build a platform providing an optimal direction. Thus, resulting in reduced amount of material and printing time. This software accompanies the 3D printers at no extra cost and usually possesses a simple graphic user interface. Figure \ref{fig: sfig01} shows the 3D model designed on Rhinoceros 3D. Figure \ref{fig: sfig02} demonstrates the exported STL file from Rhinoceros 3D and imported on Simplify 3D. Figure \ref{fig: sfig03} shows the preview mode of the internal structure of 3D model. After performing these steps the print command can be sent to the printer and the finished product can be obtained as can be seen in  Figure \ref{fig: sfig04}.

\section{methods and material}
\label{Sec4}
\subsection{3D Printer and its Working Principle}
\begin{figure}[htbp]
\begin{center}
\includegraphics[scale=0.4]{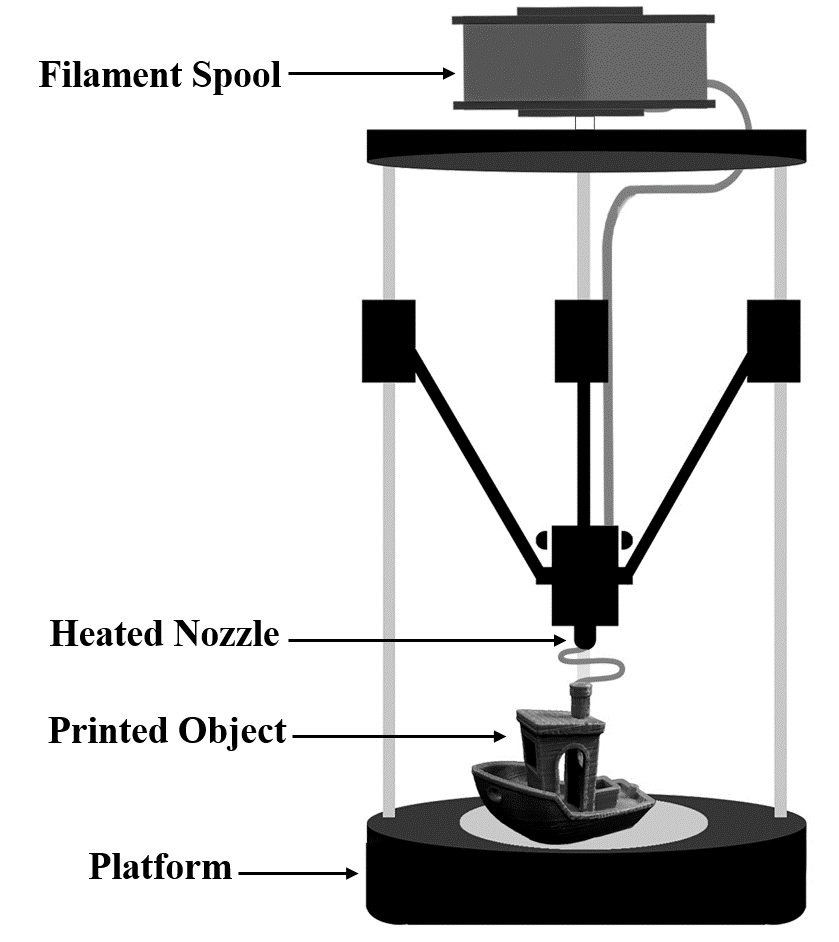}
\end{center}
\caption{ Delta 3D Printer}
\label{fig3}
\end{figure}

 For experimental purpose, Delta 3D printer is used which is capable of building small design i.e (180x180x320)mm. This 3D printer has the minimum printing speed 80mm/sec and maximum printing speed 150mm/sec, resolution 50-200 microns, and net weight of 10 kg. \cite{websitecite3} \cite{websitecite4}. This 3D printer takes the 3D model in the form of STL file. The material provided to the printer is plastic filament spool commonly known as polyactic acid (PLA) plastic. This filament is loaded with the help of a motor and is melted in nozzle section at 180-200$^{\circ}$C and renders the printed 3D objects on platform. The elements of the printer are shown in Figure \ref{fig3}.

\subsection{Estimating the Printing Speed of 3D Printer}
\begin{figure}[h!]
	\begin{center}
		\centering
		\includegraphics[width=8cm]{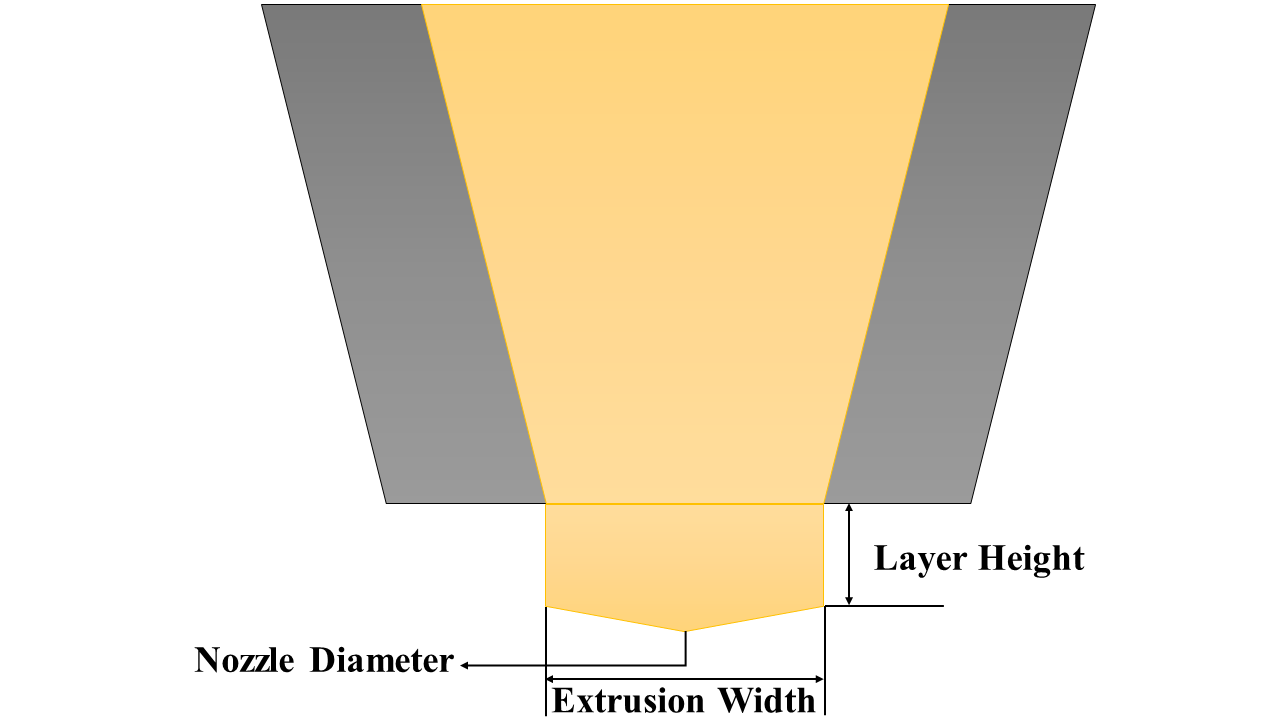}
		\end{center}
	\caption{Section View of Nozzle}
	\label{fig4}
\end{figure}

Choosing the actual speed of 3D printer can be a challenge for new user of 3D printing. Hardware must not be configured using random values. After some  hit and trials, one can estimate the speed of a slicer software that works best for the printer under test. The extrusion speed depends upon two parameters i.e layer height and extrusion width. To find the speed of the extrusion, equations \ref{eq1}, \ref{eq2} and \ref{eq3} are proposed for estimating the printing speed of 3D printer in \cite{websitecite} \cite{websitecite1}. The parameters of the equations are related to the section view of nozzle which is shown in Figure \ref{fig4}.

\begin{equation}
Print \ Area = Extrusion \ Width * Layer \ Height
\label{eq1}
\end{equation}

\begin{equation}
Melt \ Volume = Print \ Speed * Print \ Area
\label{eq2}
\end{equation}

\begin{equation}
Maximum \ Print \ Speed =\dfrac{Melt \ Volume}{Print \ Area}
\label{eq3}
\end{equation}

\section{results}
\label{Sec5}
\begin{figure}[htbp] 
\begin{center}
\begin{subfigure}{.23\textwidth}
\centering
\includegraphics[width=0.87\linewidth]{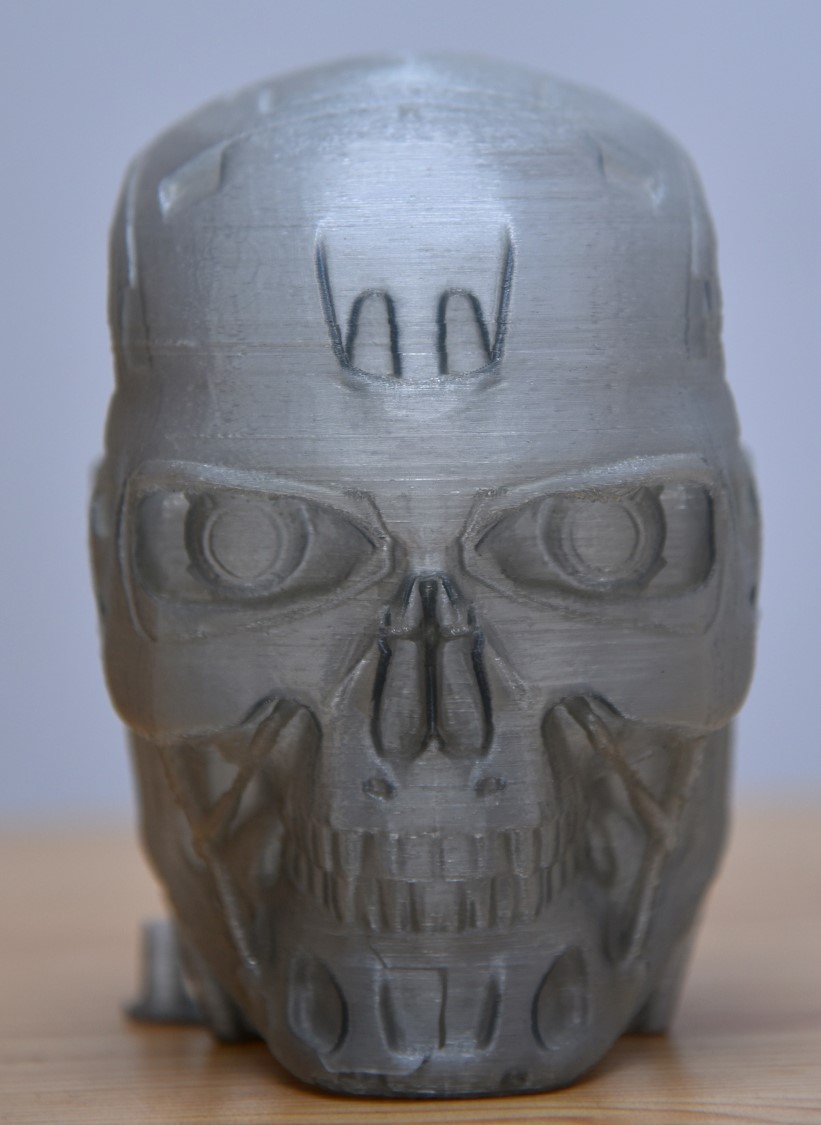}
\caption{Skull}
\label{fig: sfig1}
\end{subfigure} 
\quad 
\begin{subfigure}{.23\textwidth}
\centering
\includegraphics[width=0.87\linewidth]{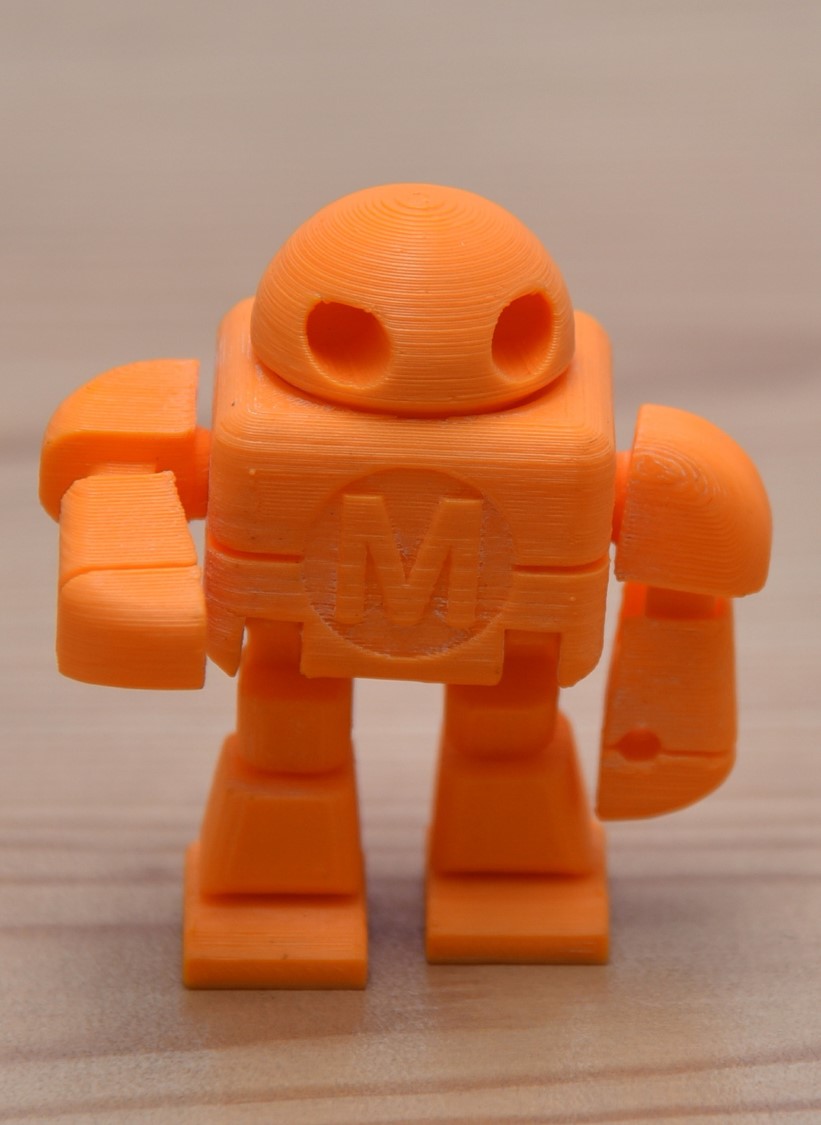}
\caption{Robot}
\label{fig: sfig2}
\end{subfigure}

\begin{subfigure}{.23\textwidth}
\centering
\includegraphics[width=0.87\linewidth]{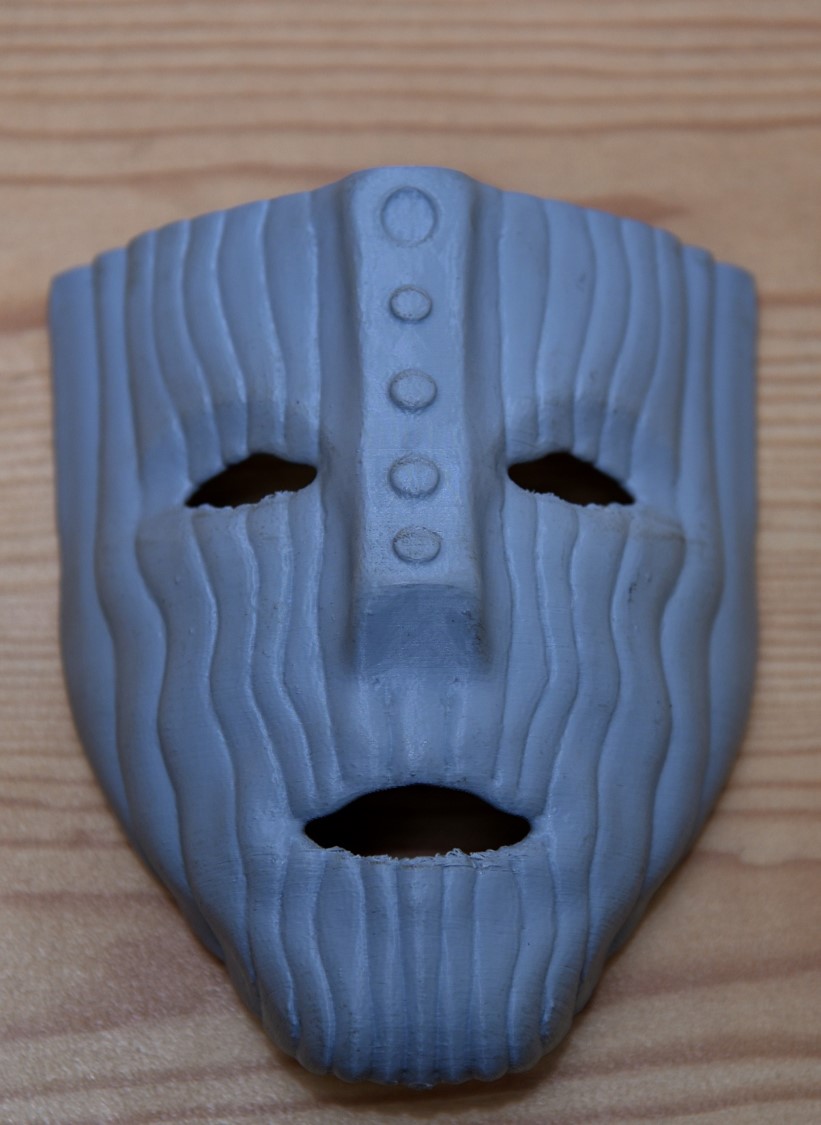}
\caption{Mask}
\label{fig: sfig3}
\end{subfigure} 
\quad 
\begin{subfigure}{.23\textwidth}
\centering
\includegraphics[width=0.87\linewidth]{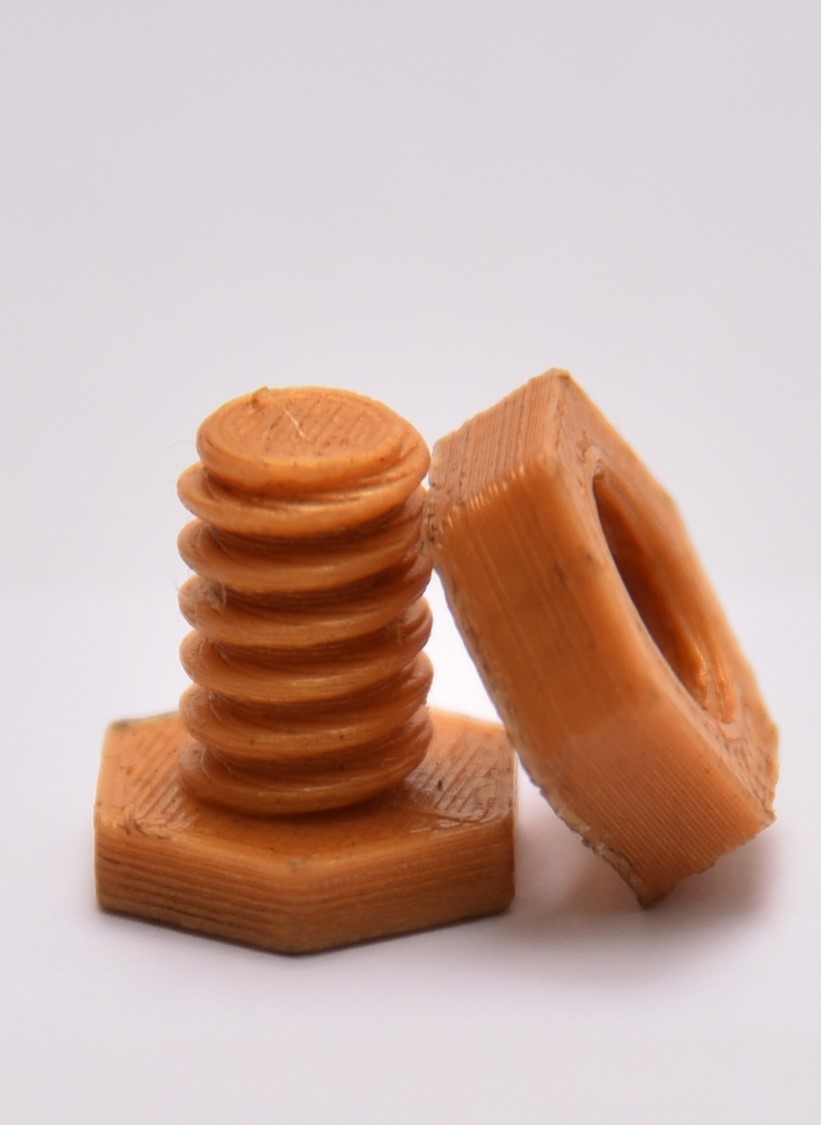}
\caption{Nut-bolt}
\label{fig: sfig4}
\end{subfigure}

\begin{subfigure}{.23\textwidth}
\centering
\includegraphics[width=0.87\linewidth]{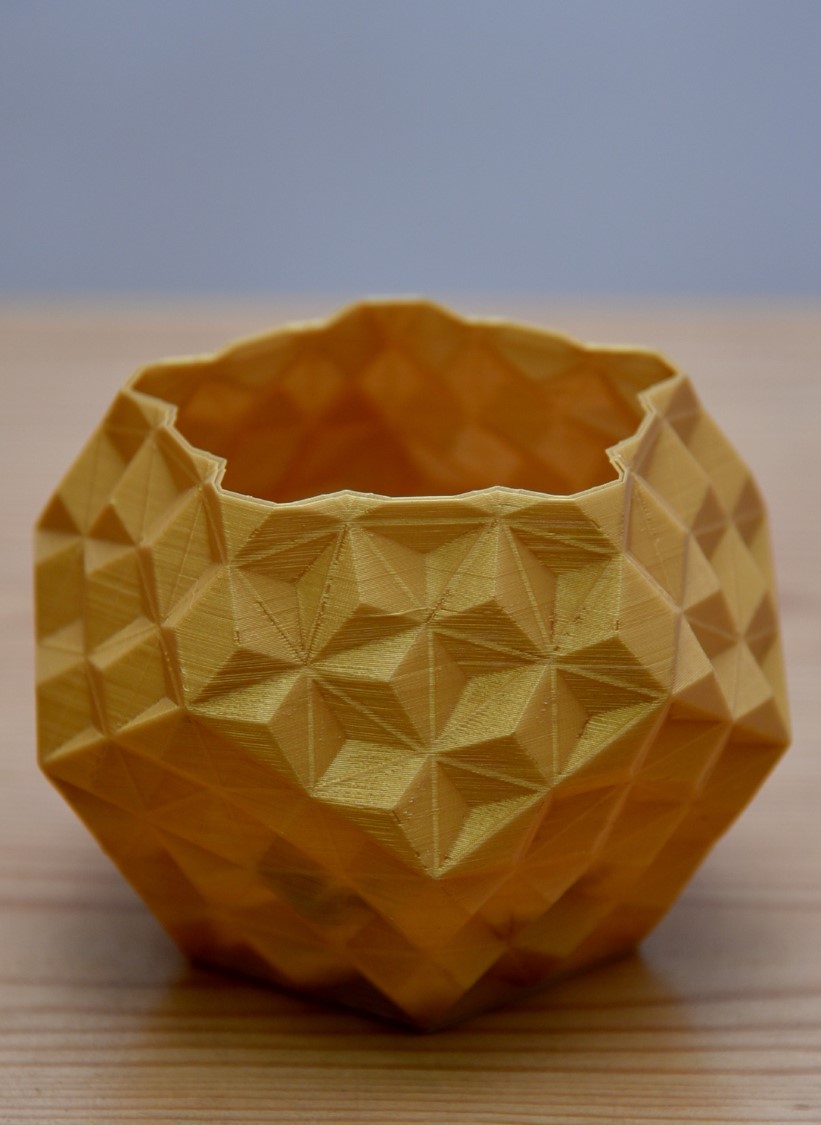}
\caption{Vase}
\label{fig: sfig5}
\end{subfigure}
\quad 
\begin{subfigure}{.23\textwidth}
\centering
\includegraphics[width=0.87\linewidth]{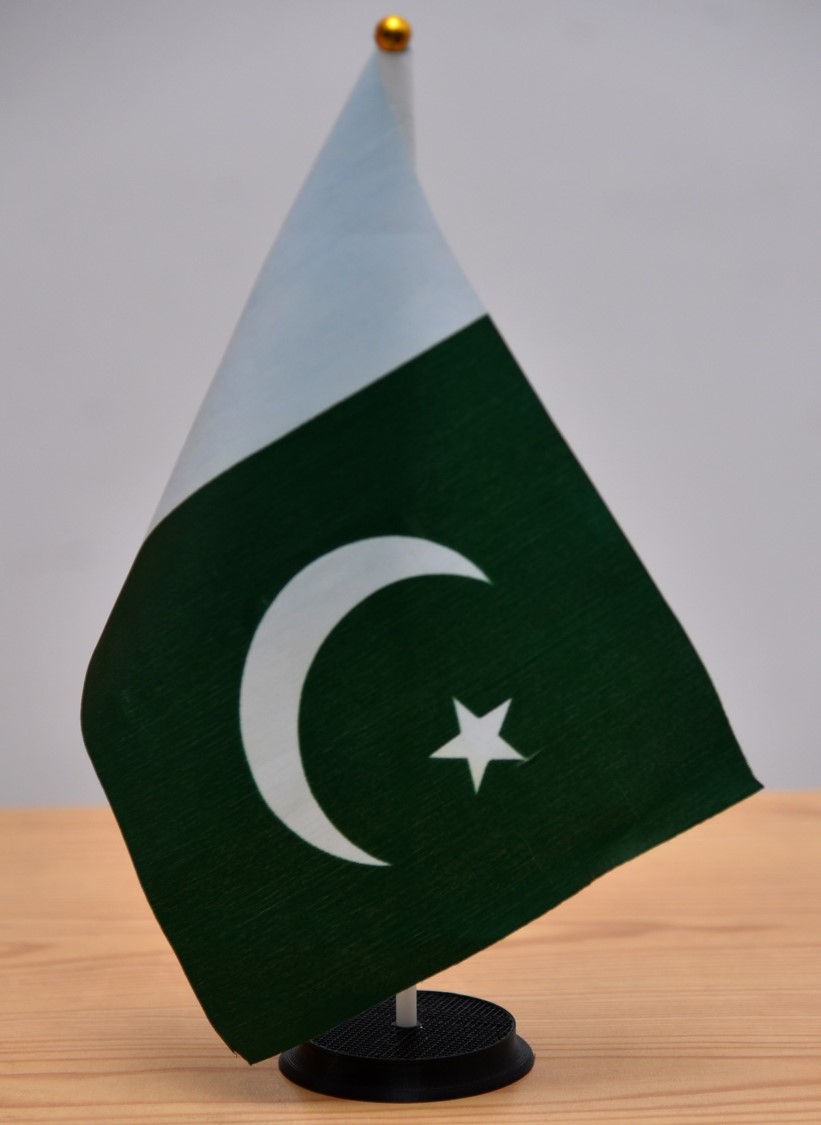}
\caption{Flag Stand}
\label{fig: sfig6}
\end{subfigure}

\begin{subfigure}{.23\textwidth}
\centering
\includegraphics[width=0.87\linewidth]{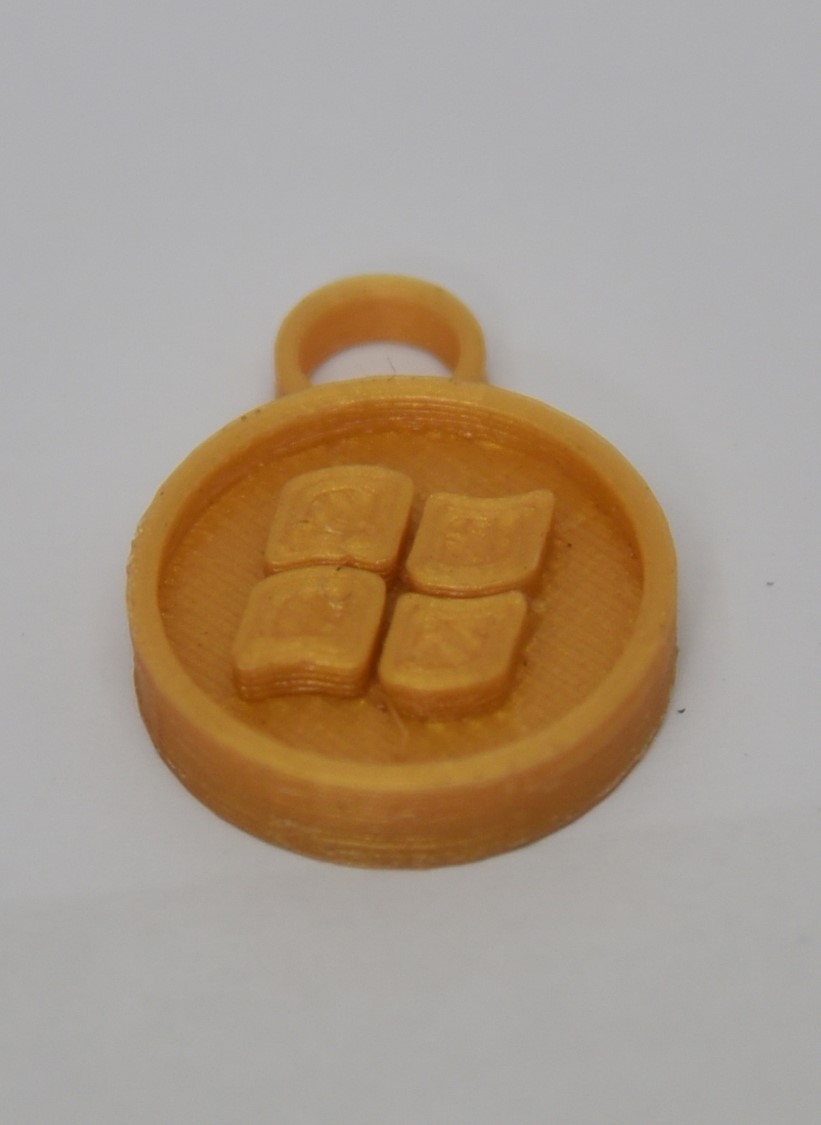}
\caption{Keyring}
\label{fig: sfig7}
\end{subfigure} 
\quad 
\begin{subfigure}{.23\textwidth}
\centering
\includegraphics[width=0.87\linewidth]{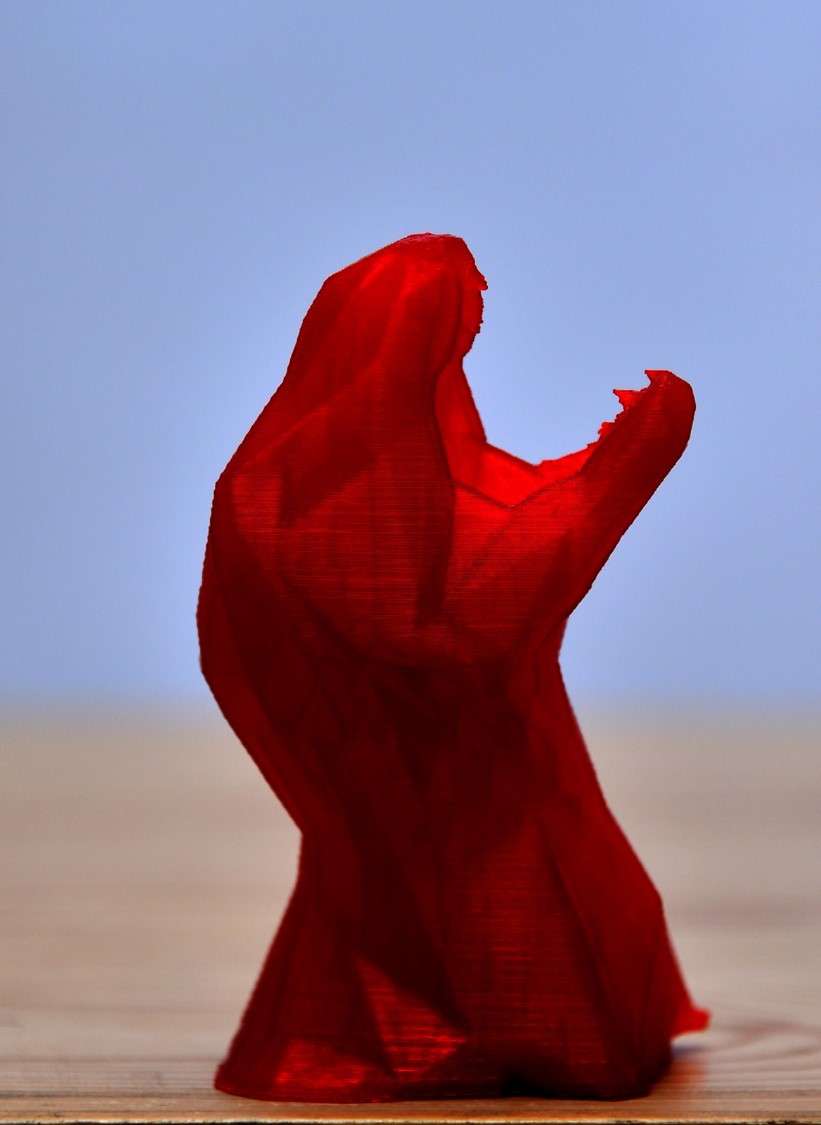}
\caption{Paper Weight}
\label{fig: sfig8}
\end{subfigure}
\caption{3D Printed Products}
\label{fig:fig}
\end{center}
\end{figure}

In this section, some 3D printed products modeled on Rhinoceros 3D and printed by using Simplify 3D as discussed in Section \ref{Sec3} are presented. Figure \ref{fig: sfig1}, \ref{fig: sfig2} and \ref{fig: sfig3} shows the skull, robot and mask respectively. These complex designs with extreme precision and finishing shows the utilization of 3D printing in manufacturing of robotics or medical products. Figure \ref{fig: sfig4} shows the nut-bolt that can be utilized as a prototype for industrial and automotive products. Figure \ref{fig: sfig5}, \ref{fig: sfig6}, \ref{fig: sfig7} and \ref{fig: sfig8} shows the vase, flag stand, keyring and paper weight respectively. This shows that the decorative product manufacturing industry can make efficient use of the additive manufacturing and meeting the clients need to deliver customized products. The detail specifications of each product are listed in Table \ref{table1}.

\begin{table}[htbp]
\caption{Description of 3D Printed Products}
\label{table1}
\begin{center}
\begin{tabular}{|c|c|c|c|}
\hline
\textbf{Product} & \textbf{\makecell{Dimension\\(mm)}} & \textbf{\makecell{Material Usage\\ (gm)}} & \textbf{\makecell{Built Time \\(min)}}\\
\hline
Skull & 130x101x148 & 413 & 1776  \\ \hline
Robot & 32x30x74 & 56 & 176 \\ \hline
Mask & 120x69x150 & 113 & 579 \\ \hline
Nut-bold & 68x31x25 & 8 & 56 \\ \hline
Vase & 70x70x90 & 87 & 517 \\ \hline
Flag Stand & 70x69x26 & 36 & 159 \\ \hline 
Keyring & 20x27x5 & 1.5 & 10 \\ \hline
Paper Weight & 60x50x120 & 67 & 196 \\ 

\hline
\multicolumn{4}{l}{mm: millimeter; gm: grams; min: minutes} 
\end{tabular}
\label{tab1}
\end{center}
\end{table}

\section{Comparison with traditional manufacturing}
\label{Sec6}
\begin{figure}[ht]
\begin{center}
\includegraphics[width=8cm]{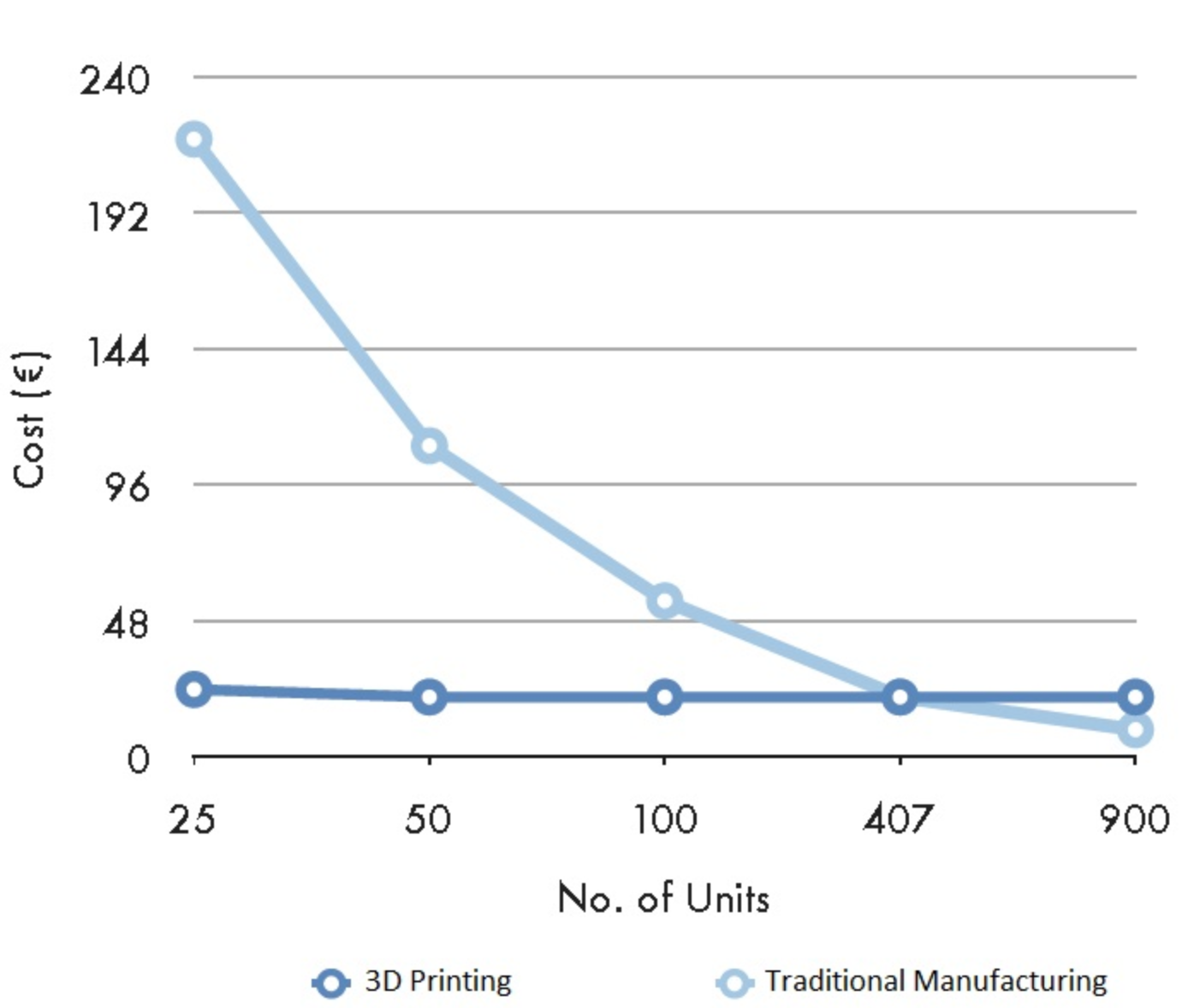}
\end{center}
\caption{3D Printing Vs Traditional Manufacturing}
\label{fig7}
\end{figure}

3D printing allows production cost to stay the same regardless of the numbers of units. Therefore, it has many implications. It allows production on small business which eliminates the need to have excess funding. Also, it reduces lead time for short production and allows each piece to be customized and unique. Additive manufacturing makes complex shapes without additional cost where as traditional manufacturing like injection molding, computer numerical control (CNC) machining  requires mass production, labor for assembly and production unit costs thousands of dollar \cite{websitecite5}. Therefore, traditional manufacturing is not suitable for prototyping. 3D printer works similar to inkjet printers or CNC machining, but instead of using multicolored inks or metal sheets, filament material like polyactic acid (PLA) plastic or acrylonitrile butadiene styrene (ABS) are used that are fed to the 3D printer. 3D printers can generate simple object in less than one hour. They can also produce objects with free moving parts that don’t need to be assembled \cite{berman20123}. The comparison between 3D printing and traditional manufacturing is shown in Table \ref{table2}. Figure \ref{fig7} shows the comparison of cost producing through conventional method vs using 3D printer.

\begin{table}[htbp]
\caption{Comparison between 3D Printing and Traditional Manufacturing}
\label{table2}
\begin{center}
\begin{tabular}{|c|c|}
\hline
\textbf{3D Printing} & \textbf{Traditional Manufacturing}\\
\hline
Short production works & Long production works  \\ \hline
Prototyping and reduced waste & Creation of perfect finish products \\ \hline
\makecell{Customizing and availability \\ of different materials} & \makecell{Unavailability of different \\materials} \\ \hline
Build complex design & Handle more products variants \\ \hline
Short time to market & Long time to market \\

\hline
\end{tabular}
\label{tab1}
\end{center}
\end{table}

\section{Conclusion}
\label{Sec7}
In this paper, we presented applications of 3D printing and several 3D printed products for general and specific fields. Low cost solution and short time to market is the major factor applicable in engineering field. Rhinoceros 3D modeling software is used for designing purpose and simulation software Simplify 3D is used for verifying and viewing 3D printing processes. A comparison of 3D modeling and slicing software, 3D printer features and parameters,
optimizing speed of extrusion, experimental results and comparison with traditional manufacturing is presented. 
As majority of related technology focus mostly on technology advancement, this study can raise the awareness of application of 3D printing. This study may also provide new directions for improving 3D printing technology to make it easier, better and more accessible in the future.

\bibliographystyle{IEEEtran}
\bibliography{ref1}

\end{document}